\documentclass[a4paper,conference]{IEEEtran}

\usepackage{times}
\usepackage{epsfig}
\usepackage{graphicx}
\usepackage{amsmath}
\usepackage{amssymb}
\usepackage[utf8]{inputenc}
\usepackage{pdfpages}
\usepackage{caption}
\usepackage{multirow}
\usepackage{url}
\usepackage[breaklinks=true,bookmarks=false]{hyperref}

\begin{document}

\title{On the Comparison of Classic and Deep\\Keypoint Detector and Descriptor Methods}

\author{
\IEEEauthorblockN{
  David Bojani\'c\IEEEauthorrefmark{1},
  Kristijan Bartol\IEEEauthorrefmark{1},\\
  Tomislav Pribani\'c,
  Tomislav Petkovi\'c
}
\IEEEauthorblockA{
  University of Zagreb\\
  Faculty of Electrical Engineering and Computing\\
  Zagreb, Croatia, EU\\
  \{name.surname\}@fer.hr\\
  \IEEEauthorrefmark{1}Equal contribution.
}
\and
\IEEEauthorblockN{
  Yago Diez Donoso
}
\IEEEauthorblockA{
  Yamagata University\\
  Faculty of Science\\
  Yamagata, Japan\\
  yago@sci.kj.yamagata-u.ac.jp
}
\and
\IEEEauthorblockN{
  Joaquim Salvi Mas
}
\IEEEauthorblockA{
  University of Girona\\
  Department of Computer Architecture\\
  and Technology\\
  Girona, Spain, EU\\
  joaquim.salvi@udg.edu
}
}

\maketitle

\begin{abstract}
The purpose of this study is to give a performance comparison between several classic hand-crafted and deep keypoint detector and descriptor methods. In particular, we consider the following classical algorithms: SIFT, SURF, ORB, FAST, BRISK, MSER, HARRIS, KAZE, AKAZE, AGAST, GFTT, FREAK, BRIEF and RootSIFT, where a subset of all combinations is paired into detector-descriptor pipelines. Additionally, we analyze the performance of two recent and perspective deep detector-descriptor models, LF-Net and SuperPoint. Our benchmark relies on the HPSequences dataset that provides real and diverse images under various geometric and illumination changes. We analyze the performance on three evaluation tasks: keypoint verification, image matching and keypoint retrieval. The results show that certain classic and deep approaches are still comparable, with some classic detector-descriptor combinations overperforming pretrained deep models. In terms of the execution times of tested implementations, SuperPoint model is the fastest, followed by ORB. The source code is published on \url{https://github.com/kristijanbartol/keypoint-algorithms-benchmark}.
\end{abstract}


\section{Introduction}

The task of keypoint detection and description has been an active area of research for a long time. An image keypoint or feature can be described as a specific meaningful structure in that image, but it is semantically ill-defined, i.e., it is not clear what are the relevant keypoints for an arbitrary input image \cite{comparative, superpoint}. The critical expectation of a good keypoint extraction method is to provide a geometric and a photometric invariance. The former assumes an invariance to image translation, rotation and scale while the latter should assure an invariance to changes in brightness, contrast and color. Keypoint detection and description methods are used as a basis for more complex computer vision tasks such as object recognition \cite{sift}, structure from motion (SfM) \cite{sfm}, simultaneous localization and mapping (SLAM) \cite{deepslam}, 3D reconstruction \cite{3d-eval}, image matching \cite{eval-mikolajczyk} and content-based retrieval \cite{hpatches}; the performance of keypoint detectors and descriptors is evaluated on these tasks.

In most cases, the practical usage of keypoint detectors assumes a process of keypoint matching across images. The matching is based on the direct utilization of an arbitrarily defined similarity measure, like L2-norm or Hamming distance \cite{hpatches, sift}. The first decade of this century has been dominated by works on SIFT-like detectors and descriptors \cite{sift, surf} which rely on distinctive blobs and corners. Another significant group of classical algorithms are binary descriptors, namely ORB \cite{orb}, BRIEF \cite{brief} and BRISK \cite{brisk}, that are generally more time-efficient than SIFT as was shown in previous benchmarks \cite{hpatches, comparative, fast-eval}.

With the rise of deep learning methods, in particular convolutional models, new keypoint detection and description approaches emerged \cite{superpoint, lfnet} claiming superior results on benchmarks over the classical algorithms. The expectation for deep models is to learn abstract image features from high-dimensional data \cite{cnn-insight}. Instead of hand crafting the features, convolutional models learn about them based on supervision, thus, their performance heavily relies on ground truth information. 

An unbiased evalution over state-of-the-art keypoint detection and description methods on multiple higher level tasks, like image matching or object detection, is crucial. Our work is partly motivated by the proposed descriptor evaluation benchmark by Balntas et al. \cite{hpatches} that shows a dominance of deep descriptors in terms of average precision. On the other hand, a keypoint detector benchmark by Lenc et al. \cite{large-scale-eval} published afterward shows that the classical detectors are still relevant in image retrieval. The aim of this work is therefore to gather these insights by performing an evaluation over joint detector and descriptor pipelines based on three evaluation tasks: keypoint verification, image matching and keypoint retrieval and on HPSequences dataset, all introduced in \cite{hpatches}. We show that some combinations of classic detectors and descriptors are indeed relevant on keypoint verification, retrieval and image matching tasks over various viewpoint and illumination changes. We also briefly summarize the most representative classic detectors and descriptors introduced both in the distant and in the more recent past. LF-Net \cite{lfnet} and SuperPoint \cite{superpoint} are chosen as deep learning representatives.

The remainder of this paper is structured as follows. In the second section we provide a concise overview of the related work. We then describe the benchmark pipeline and comment on the results. Finally, the concluding remarks are provided. 

\section{Related Work} \label{related}

We first give a literature review in order of detector-only, descriptor-only and detector-descriptor algorithms respectively; distinguishing between classical and deep models.
Next, we give a brief overview of relevant benchmarks, where we highlight six studies that introduce novel algorithm review methods.

\subsection{Detectors and Descriptors}
HARRIS \cite{harris} detector measures the similarity of an image patch centered around a point with overlapping nearby patches using the sum-of-squared-differences of the pixel intensities. The measure can be expressed using the eigenvalues of the sum-of-squared-differences matrix obtained by the Taylor expansion of the initial measure. GFTT \cite{gftt} improves on this by simplifying the self-similarity measure using only the minima of the mentioned eigenvalues. With the goal of speeding up keypoint retrieval, FAST \cite{fast} tests a number of brighter or darker pixels in a ring around a given point, after which it uses a decision tree classifier that was previously learned on a set of similar images to improve efficiency.
AGAST \cite{agast} aims to speed up FAST by using a generic binary decision tree which does not have to be adapted to new environments.
MSER \cite{mser} binarizes the image at multiple thresholds, and finds stable regions that do not significantly change over a number of thresholds.

Two representatives of descriptor-only approaches are both binary descriptors: BRIEF \cite{brief} and FREAK \cite{freak}. BRIEF randomly selects pairs of points around a keypoint to  create a binary descriptor using pixel intensity comparisons between the selected pairs. Similarly, FREAK uses a circular retinal sampling grid that has a higher density of points near the center, and compares pixel intensities to create a binary descriptor.

One of the most prominent detector and descriptor algorithms is SIFT \cite{sift}. As a detector, SIFT convolves the image with Gaussian filters at various scales and detector finds scale invariant keypoints by selecting the local extrema in both scale and space. The SIFT descriptor is a histogram of local image gradient directions around the interest point. RootSIFT \cite{rootsift} differentiates from SIFT by using a Hellinger distance instead of the standard Euclidean distance to measure the similarity between its descriptors. SURF \cite{surf} aims to speed up SIFT by convolving the image with an approximation of the second-order derivative of the Gaussian filter using integral images. The SURF descriptor assigns the orientation to a keypoint by calculating the Haar wavelet response in the $x$ and $y$ directions of a circular neighbourhood.
ORB \cite{orb} (Oriented FAST and Rotated BRIEF) tries to further speed up SIFT and SURF. The ORB detector uses FAST to compute the keypoints and adds rotation invariance by assigning them an orientation by the intensity weighted centroid. The ORB descriptor relies on BRIEF and adds rotation invariance by rotating the point pairs with the previously found keypoint orientation. 
BRISK \cite{brisk} identifies points of interest in the image pyramid using a saliency criterion. For every keypoint, the orientation is obtained by applying a sampling pattern to its neighbourhood, and retrieving gray values.
Among newer classical algorithms, we distinguish KAZE \cite{kaze} and AKAZE \cite{akaze}. KAZE detector is based on a scale normalized determinant of the Hessian matrix which is computed at multiple scale levels. The maxima of the detector responses are picked up as keypoints using a moving window. The rotation invariant KAZE descriptor is obtained by finding the dominant orientation in a circular neighbourhood around each keypoint. AKAZE accelerates KAZE by using a more computationally efficient framework.

In contrast to the classical algorithms, deep models learn to detect and describe local keypoints in an end-to-end fashion from scratch. The inference is done in a single forward propagation step. LF-Net \cite{lfnet} detector generates a scale-space keypoint detections along with dense orientation estimates, which are then used to select a fixed number of 512 keypoints. Image patches around the selected keypoints are cropped and fed into a descriptor network. The output is a set of keypoint descriptors. In the learning phase, LF-Net is a two-branch architecture. The first branch generates supervision for the second branch using the ground truth obtained from conventional SfM methods. SuperPoint \cite{superpoint} is an encoder-decoder architecture, where encoder branches into two decoders, one for the detector and one for the descriptor. SuperPoint consists of a three-step learning pipeline. In the first step, the model learns to detect keypoints based on supervision from a synthetic dataset, where keypoints can be determined unambiguously. In the second step, the detector capacity is expanded to real images using homographic adaption. Homographic adaptation is designed to enable self-supervised training. Finally, a keypoint descriptor is trained on image matching task based on labels obtained from self-supervision.

\subsection{Evaluation Benchmarks}

The first complete, large-scale  and still popular keypoint detector evaluation benchmark which proposes specific evaluation metrics and methods was created by Mikolajczyk et al. \cite{affine-det-comp}. The authors discuss base image detection terminology and argue that, for viewpoint changes, the affine transformation is of most interest. Photometric transformations can be modeled by a linear transformation of the intensities. 

A comprehensive experimental study by Mukherjee et al. \cite{comparative} evaluates a vast amount of classical detector and descriptor approaches. Some of the benchmark algorithms are detector-only \cite{harris, fast, mser, agast, gftt}, some are descriptor-only \cite{brief, rootsift, freak}, and some are both \cite{sift, surf, orb, brisk, akaze, kaze}. From the set of detector and descriptor combinations, they comment on the results of the most successful ones. The study by Mukherjee et al. \cite{comparative} even proposes detector and descriptor evaluation metrics, but it lacks a benchmark over deep learning approaches.

A very important keypoint detector metric is repeatability which measures detector's ability to identify the same features despite geometric and illumination changes. A critique on the bias in the repeatability criterion of previous benchmarks was stated by Rey-Otero et al. \cite{rep-bias}. The authors argue that less selective detectors tend to score higher on repeatability as they have a greater chance of hitting the same keypoints over multiple views.

The most recent comprehensive feature detection benchmark was made by Lenc et al. \cite{large-scale-eval}, where the HPSequences dataset is also used (\autoref{fig:hpsequences}). They cope with the biased repeatability issue by simply limiting the number of features a detector can extract. The benchmark by Balntas et al. \cite{hpatches} evaluates classic and deep descriptors. To make a fair comparison, they prepare the HPatches dataset, which uses full-sized HPSequences images and a common detector to extract image patches that are then processed by the evaluated descriptors. Note that Balntas et al. \cite{hpatches} do not mention HPSequences under its name; the name was first used by Lenc et al. \cite{large-scale-eval}. Finally, a benchmark over both classic and deep detectors and descriptors was created by Fan et al. \cite{3d-eval}, but solely based on a 3D reconstruction task.

\section{Evaluation Setup} \label{experimental}

The main goal of this work is to compare classical and deep learning detector-descriptor pipelines. To measure the performance of the whole pipeline, we use three evaluation tasks: keypoint verification, image matching and keypoint retrieval inspired by the descriptor evaluation benchmark by Balntas et al. \cite{hpatches}. In contrast to \cite{hpatches}, that uses predefined patches, we use the original image sequences called HPSequences, because keypoints need to be obtained again to evaluate every detector-descriptor combination properly.

\subsection{HPSequences Dataset}

\begin{figure}[h]
  \includegraphics[height=0.217\textwidth]{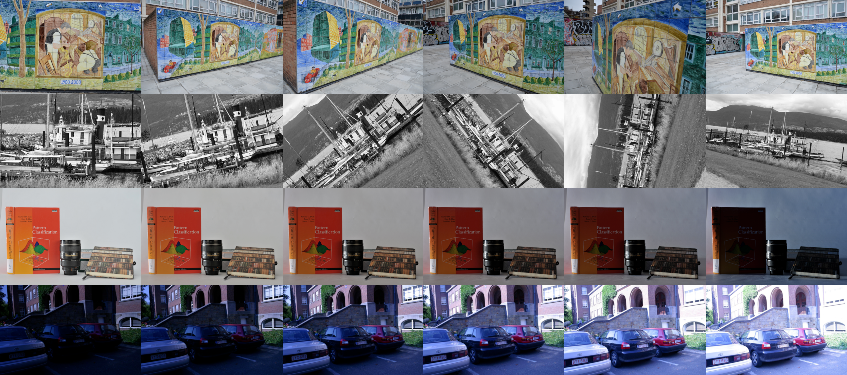}
  \caption{Examples from the \textit{HPSequences} dataset \cite{hpatches} displaying two viewpoint (top) and two illumination (bottom) based sequences.}
  \label{fig:hpsequences}
\end{figure}

HPSequences dataset is a collection of image sequences gathered by Balntas et al. \cite{hpatches}. The dataset consists of $116$ image sequences, $57$ of which represent only photometric changes, whereas $59$ represent only geometric deformations. We denote these two groups of sequences as \textit{illumination} and \textit{viewpoint} changes, respectively. In \autoref{fig:hpsequences}, first two sequences show viewpoint and the other two show illumination changes.

Each sequence consists of one reference image and $5$ target images representing the appropriate illumination or viewpoint changes. Alongisde every target image there is a homography connecting it to the reference image. In case of an illumination change sequence, the homography is an identity mapping. The HPSequences dataset is significant as it is real, diverse and large enough to be used as a standalone dataset for the evaluation benchmark.

\subsection{Evaluation Pipeline}

\begin{table*}[t]
\begin{tabular}{|l|c|c|c||l|c|c|c||l|c|c|c|}
\hline
Nr. & Detector & Descriptor & Time(ms) & Nr. & Detector & Descriptor & Time(ms) & Nr. & Detector & Descriptor & Time(ms) \\
\hline
1 & SUPERPOINT & SUPERPOINT & 13 & 16 & SURF & BRISK & 91 & 31 & SIFT & FREAK & 147 \\
\hline
2 & ORB & ORB & 17 & 17 & HARRIS & BRISK & 92 & 32 & FAST & SIFT & 163 \\
\hline
3 & GFTT & BRIEF & 24 & 18 & FAST & BRISK & 95 & 33 & AGAST & SIFT & 181 \\
\hline
4 & GFTT & SURF & 24 & 19 & HARRIS & FREAK & 103 & 34 & SIFT & SIFT & 195 \\
\hline
5 & GFTT & BRISK & 27 & 20 & FAST & FREAK & 104 & 35 & LFNET & LFNET & 196 \\
\hline
6 & FAST & BRIEF & 32 & 21 & BRISK & BRISK & 105 & 36 & BRISK & SIFT & 204 \\
\hline
7 & GFTT & SIFT & 36 & 22 & SURF & FREAK & 111 & 37 & SIFT & ROOT\textunderscore SIFT & 207 \\
\hline
8 & HARRIS & SIFT & 39 & 23 & AGAST & BRISK & 117 & 38 & SURF & SURF & 208 \\
\hline
9 & HARRIS & BRIEF & 43 & 24 & AKAZE & AKAZE & 118 & 39 & MSER & BRIEF & 209 \\
\hline
10 & AGAST & BRIEF & 46 & 25 & SIFT & BRIEF & 122 & 40 & MSER & BRISK & 214 \\
\hline
11 & GFTT & FREAK & 57 & 26 & SIFT & BRISK & 125 & 41 & MSER & SURF & 228 \\
\hline
12 & FAST & SURF & 57 & 27 & AGAST & FREAK & 126 & 42 & MSER & FREAK & 242 \\
\hline
13 & SURF & BRIEF & 66 & 28 & BRISK & FREAK & 130 & 43 & MSER & SIFT & 524 \\
\hline
14 & AGAST & SURF & 81 & 29 & SIFT & SURF & 131 & 44 & SURF & SIFT & 555 \\
\hline
15 & BRISK & BRIEF & 81 & 30 & BRISK & SURF & 146 & 45 & KAZE & KAZE & 716 \\
\hline
\end{tabular}
\centering
\caption{\label{tab:exe-time}Mean execution times in milliseconds for all the 45 detector-descriptor pairs sorted in the asceding order.}
\end{table*}

Our evaluation relies on Balntas et al. \cite{hpatches} with the difference of evaluating both detectors and descriptors.
The evaluation tasks, namely keypoint verification, image matching and keypoint retrieval, make use of the average precision measure.

The average precision (AP) is a measure based on precision and recall of a ranked list $L_K$ with $K$ elements. For every $k < K$ we compute the precision and recall for $L_k$, where $k$ indicates the top-$k$ elements of the ranked list. By averaging the precisions computed for every $L_k$ for which the recall increases, we obtain the AP measure of the ranked list $L_K$.

We begin by finding keypoints and descriptors for the selected pairs of detectors and descriptors shown in \autoref{fig:results}.
Let $I_{ij}$ denote the $j$-th image from the $i$-th sequence and $K_{ij}$ the set of obtained keypoints for that image, where $j=1,\ldots,6$ and $i=1,\ldots,116$. As a convention, the first image $I_{i1}$ is the reference image. Let $H_{ij}$ denote the homography from the reference image to the $j$-th image in sequece $i$. $H_{i1}$ is an identity mapping.
Let $K_{i1}' = \left\{ x_k \mid k=1,\ldots,n\right\} $ be a set of $n$ random keypoints from the reference image for a sequence $i$.

\vspace{0.2cm}
\noindent \textbf{(a) Keypoint verification}. For a keypoint $x_k \in K_{i1}'$ we find the matching keypoints $x_{ij,k}'$ from images $I_{ij}, j=2,\ldots,6$, as well as from a random selection of images $I_{pl}$ not belonging to the $i$-th sequence ( $p\neq i$ and $l\in \left\{ 1,\ldots,6\right\}$). The matches are found using the minimal descriptor distance between keypoints. This results in a set
\begin{equation}
\label{eq:verification-set-A}
    A_i = \left\{ \left(x_k, x_{ij,k}', s_{ij,k}, y_{ij,k}\right), k \in \mathbb{N}\right\},
\end{equation}
where $s_{ij,k}$ is the descriptor distance (being Euclidean or Hamming, depending on the descriptor type) between $x_k$ and $x_{ij,k}'$ and where
\begin{equation}
\label{eq:pV-y}
y_{ij,k} = \left\{\begin{array}{@{}ll@{}}
        \multirow{2}{*}{+1, } & \text{if } d(H_{ij}x_k,x_{ij,k}') \leq                              d(H_{ij}x_k,x') \\
                              & \quad \forall x' \in K_{ij} \text{ and } x_k,x' \text{in sequence } i,\\
        -1, & \text{otherwise},
        \end{array}\right.
\end{equation}
is the appropriate label and $d$ is the Euclidean distance. In other words, $y_{ij,k}$ is $+1$ if $x_{ij,k}'$ is the closest keypoint from $K_{ij}$ to the homography projection of $x_k$ on image $j$.
Repeating the process for every keypoint from $K_{i1}'$ results in a set \\
$A = \cup_{i=1}^n A_i$ which is used to evaluate the given detector-descriptor pair using the AP measure.

\vspace{0.2cm}
\noindent \textbf{(b) Image matching}. We match the keypoints from $K_{i1}'$ with keypoints from $K_{ij}$ in image $I_{ij}$, $j=2,\ldots,6$. Matching is performed by finding the minimal distance between the keypoints descriptors from one image to another.
The matching between images $I_{i1}$ and $I_{ij}$, $j\neq 1$, results in a set 
\begin{equation}
\label{eq:verification-set-B}
B = \left\{ \left( x,x',s,y \right) \mid x\in K_{i1}', x' \in K_{ij} \right\},
\end{equation}
where $s$ is the distance between $x$ and $x'$ descriptors (Euclidean or Hamming, depending on the descriptor type) and where
\begin{equation}
\label{eq:iM-y}
y = \left\{\begin{array}{ll}
        +1, & \text{if } d\left(H_{ij}x,x'\right) \leq                              d(H_{ij}x,z) \quad \forall z \in K_{ij}, \\
        -1, & \text{otherwise},
        \end{array}\right.
\end{equation}
is the appropriate label and $d$ is the Euclidean distance. In other words, $y$ is $+1$ if $x'$ is the closest keypoint to the homography of $x$.

Every image matching results in such a set $B$ which gets evaluated with the AP measure. Finally, for all the AP's computed, we find the mAP by averaging over the AP's.

\vspace{0.2cm}
\noindent \textbf{(c) Keypoint retrieval}. We match every keypoint $x \in K_{i1}'$ with keypoints from $K_{ij}$ on image $I_{ij}$, $j=2,\ldots,6$, as well as a number of random keypoints $K_{pl}$ from images $I_{pl}$ out of $i$-th sequence ( $p\neq i$ and $l\in \left\{ 1,\ldots,6\right\}$). This results in a set 
\begin{align}
\label{eq:verification-set-C}
C = \big\{ \left(x,x',s,y\right) \mid & x \in K_{i1}',\nonumber\\
& x' \in K_{ij} \cup K_{pl}, \\
& j=2,\ldots,6 ,\nonumber \\
& l \in \{2,\ldots,6\}, \nonumber\\
& p\neq i  \big\} \nonumber
\end{align}
where $s$ is the distance between descriptors of keypoints $x$ and $x'$, and where
\begin{equation}
\label{eq:pR-y}
y = \left\{\begin{array}{@{}ll@{}}
         +1, & \text{if } d\left(H_{ij}x,x'\right) \leq                              d(H_{ij}x,z)  \quad \forall z \in K_{ij}, \\
         \multirow{2}{*}{0, } & \text{if } d\left(H_{ij}x,x'\right) >                              d(H_{ij}x,z)  \quad \forall z \in K_{ij}\\
                               &  \quad \text{and } x' \in K_{ij}, j \in \left\{2,\ldots,6\right\}, \\
        -1, & \text{otherwise},
        \end{array}\right.
\end{equation}
is the appropriate label and $d$ is the Euclidean distance.

Every keypoint retrieval results in such a set $C$ which gets evaluated with the AP measure. Finally, we compute the mAP by averaging over the AP's.

The three tasks described above are the adaptions of the originally proposed descriptor-only evaluation tasks \cite{hpatches}. We propose to use the described tasks for joint detector-descriptor evaluation, because the keypoint detections strongly influence both the descriptor and the overall performance.

\section{Results and discussion}

\begin{figure*}[h!]
  \centering
  \captionsetup{justification=centering}
  \includegraphics[trim={0 2cm 0 1cm},clip, width=\textwidth,height=15cm]{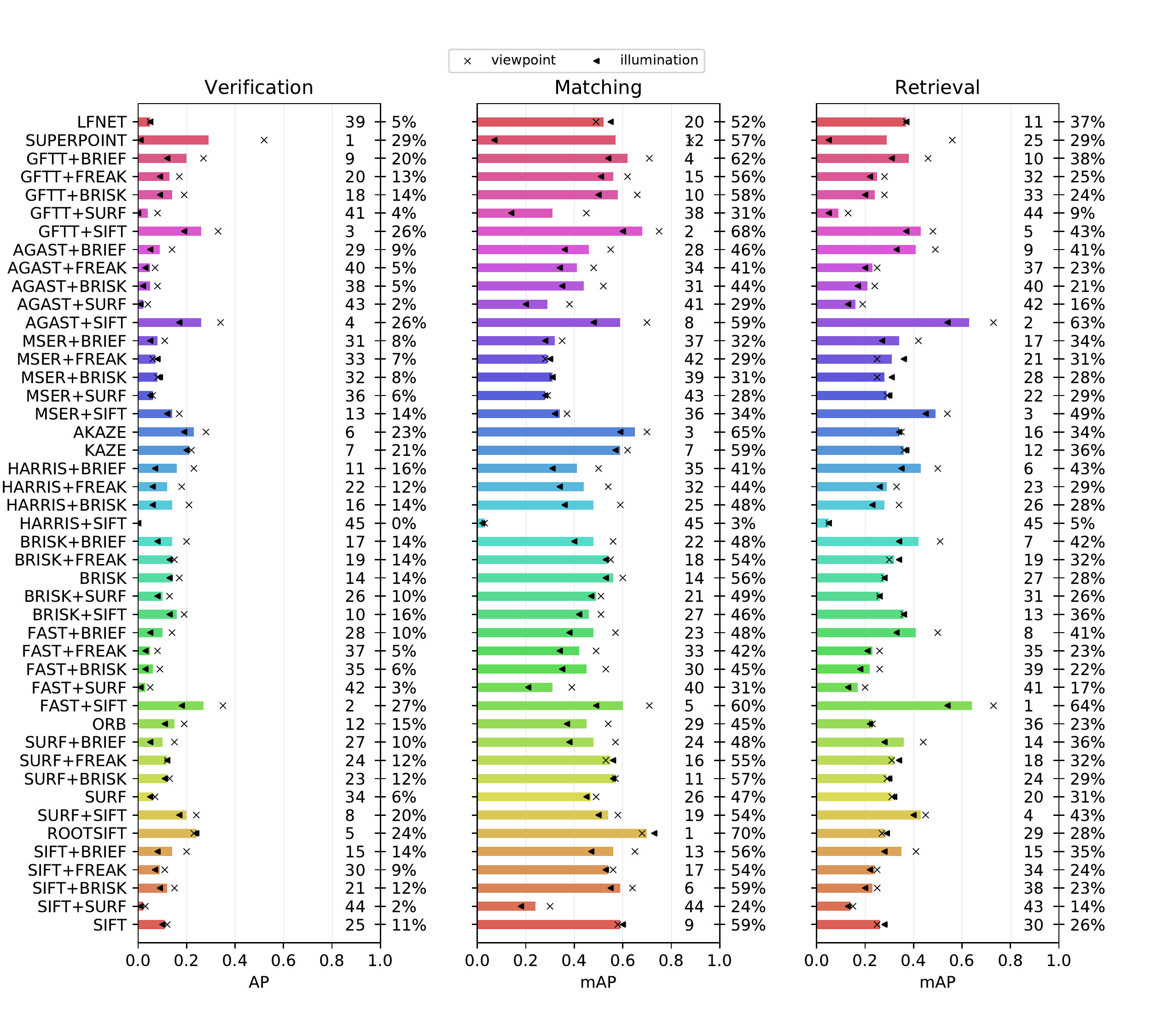}
  \caption{Experimental results for the selected detector/descriptor combinations over three evaluation methods: keypoint verification, image matching and keypoint retrieval. Combinations of different detector and descriptor algorithms are labelled as DET+DESC. As described in equation \ref{eq:verification-set-A} of section \ref{experimental}, keypoint verification selects a single set. Therefore, we only calculate AP over the set. In case of matching and retrieval, mAP is displayed. Columns display mean over both viewpoint and illumination sequences.}
  \label{fig:results}
\end{figure*}

We evaluate 45 detector-descriptor combinations in total, 43 of which are classic and 2 of which are deep algorithms. Figure \ref{fig:results} shows the experimental results for the 45 detector-descriptor combinations over three evaluation tasks under viewpoint and illumination changes. As the subset of keypoints is randomly selected for each experiment, we repeat it $m=5$ times and finally average over mAPs to get more reliable results. The median values for the three tasks are $12\%$, $48\%$ and $29\%$ for keypoint verification, image matching and keypoint retrieval, respectively. This means that the image matching task, as defined in this paper, is the easiest and the keypoint verification is the hardest evaluation task, which is in contrast to Balntas et al. \cite{hpatches}. We might explain this with the fact that keypoint verification needs to distinguish between both the positive and the negative in-sequence examples and all the negative out-of-sequence examples. Keypoint retrieval distinguishes between the positive in-sequence examples and the negative out-of-sequence examples, which is an easier task. Finally, image matching only distinguishes between the positive and the negative in-sequence examples, but using more positive examples than in the keypoint retrieval task (only 5 positives).

The dominant combinations on all three tasks are: GFTT+SIFT (3rd, 2nd and 5th) AGAST+SIFT (4th, 8th and 2nd) and FAST+SIFT (2nd, 5th and 1st), which is in correspondence with the results on the precision metric from Mukherjee et al. \cite{comparative}. Note that SIFT descriptors come up in all three most successful combinations. In contrast to SIFT, SURF detector and descriptor are shown to perform badly in all combinations (ranking 34th, 26th and 20th on the three tasks). As expected, AKAZE overperforms KAZE on 2 of 3 tasks; by $2\%$ on keypoint verification, 6$\%$ on image matching and is overperformed by KAZE on keypoint retrieval by only $2\%$. In the context of binary descriptors, ORB and BRISK generally perform better than others. ORB outperforms BRISK by $2\%$ on keypoint verification, while BRISK outperforms ORB by $11\%$ on image matching and by $5\%$ on keypoint retrieval. Finally, RootSIFT performs best with $70\%$ mAP on image matching and ranks 5th on keypoint verification task.

SuperPoint and LF-Net generally perform well on the three tasks, with SuperPoint being the most successful of all on the keypoint verification. Surprisingly, LF-Net perfoms badly on the keypoint verification task, both on viewpoint and on illumination changes. It is also worth pointing out that SuperPoint has an excellent performance under the illumination changes, while it performs very badly under the geometric changes. This might be due to the fact that SuperPoint takes a relatively small number of keypoints per image (up to 20) and this might be insufficient to describe the transformations.

The majority of detector-descriptor combinations have better results under the illumination than under the geometric changes. RootSIFT, MSER+FREAK, MSER+BRISK and LF-Net are the exceptions, having slightly better overall results under the geometric changes.

In terms of speed, SuperPoint is the fastest algorithm, followed by ORB, as shown in Table \ref{tab:exe-time}. We measure the execution time as the average time needed to detect and describe an image. Note that SuperPoint is a deep algorithm and is run on a GPU. The comparison is therefore unfair as the GPU execution is faster under the assumption that the algorithm is written specially for the GPU (which it is).

\subsection{Implementation remarks}

For LF-Net \cite{lfnet} and SuperPoint \cite{superpoint} evaluation, we use pretrained models provided on their projects' websites \cite{lfnet-github, superpoint-github}. For classical algorithms, the default OpenCV \cite{opencv} parameters were used. The proposed evaluation benchmark is therefore unbiased between classical and deep approaches.

All the experiments were done on a single computer, 64GB RAM, Intel Core i7-8700K CPU, 3.7GHz. Two 12GB VRAM Titan Xp GPUs were used for the evaluation of LF-Net and SuperPoint. During the course of this work, an open source benchmark was developed and is available at \url{https://github.com/kristijanbartol/keypoint-algorithms-benchmark}. It can be used to evaluate different detectors and descriptor pairs on the HPSequences dataset.


\section{Conclusion}

In contrast to our expectations, deep models did not overperform classic ones by a large margin. Reflecting on the results from the previous detector \cite{large-scale-eval} and descriptor \cite{hpatches} benchmarks, by Lenc and Vedaldi and Balntas et al. respectively, this is not surprising. In Lenc et al. \cite{large-scale-eval} it is shown that, while deep learning helps for illumination invariance, classical detectors are still competitive in general. Having distinctive keypoints picked by the classical detector, a keypoint descriptor has a strong basis for reaching the higher overall performance. Slightly weaker deep learning performance is also due to the fact that the pretrained models, available online, were used.

For the future work, we should fine-tune pretrained LF-Net and SuperPoint models to reach their full potential. In a similar fashion, searching for the parameters of the classical algorithms, possibly better than default ones, is also desired. In terms of benchmark improvements, we could explicitly measure repeatability and selectivity of the detector. The expectations are that the greater repeatability is proportional to the lower selectivity, as stated in Rey-Otero et al. \cite{rep-bias}, which is also proportional to the better results on the three evaluation methods used in this work. 

\section*{Acknowledgement}

This work has been supported by Croatian Science Foundation under the grant number HRZZ-IP-2018-01-8118 (STEAM) and by European Regional Development Fund under the grant number KK.01.1.1.01.0009 (DATACROSS).


\end{document}